\documentclass[letterpaper, 10 pt, conference]{ieeeconf}  

\IEEEoverridecommandlockouts
\usepackage{url}
\usepackage{balance}
\usepackage{amsmath}
\usepackage{amssymb,bm}
\usepackage{amsfonts}
\usepackage{enumerate}
\usepackage{tabulary}
\usepackage{multirow}
\usepackage{cite}
\usepackage{lipsum}  
\usepackage{graphicx}
\usepackage{epstopdf}
\usepackage[misc]{ifsym}
\usepackage{color}
\usepackage{xcolor}
\usepackage{xxcolor}
\usepackage{pgf}
\usepackage{tikz}
\usepackage{float}
\usepackage{ifthen}
\usepackage{forloop}
\usepackage{listings}
\usepackage{lipsum}
\usepackage{booktabs}
\usepackage[ruled,linesnumbered]{algorithm2e}
\usepackage{verbatim}
\usepackage{hyperref}
\usepackage{makecell}
\usepackage{xfrac}

\DeclareMathOperator*{\argmin}{argmin}
\SetAlCapSkip{0.5em}

\title{\LARGE \bf
Learning-based Bias Correction for Time Difference of Arrival \\ Ultra-wideband Localization of Resource-constrained Mobile Robots
}

\author{Wenda Zhao, Jacopo Panerati, and Angela P. Schoellig
\thanks{The authors are with the \href{http://www.dynsyslab.org}{Dynamic Systems Lab}, Institute for Aerospace Studies, University of Toronto, Canada, and affiliated with the \href{https://vectorinstitute.ai/}{Vector Institute for Artificial Intelligence} in Toronto. 
E-mails:
        {\tt \{firstname.lastname\}@utoronto.ca}}}

\begin{document}
\maketitle
\thispagestyle{empty}
\pagestyle{empty}

\begin{abstract}
Accurate indoor localization is a crucial enabling technology for many robotics applications, from warehouse management to monitoring tasks. Ultra-wideband (UWB) time difference of arrival (TDOA)-based localization is a promising lightweight, low-cost solution that can scale to a large number of devices---making it especially suited for resource-constrained multi-robot applications. However, the localization accuracy of standard, commercially available UWB radios is often insufficient due to significant measurement bias and outliers. In this letter, we address these issues by proposing a robust UWB TDOA localization framework comprising of \textit{(i)} learning-based bias correction and \textit{(ii)} M-estimation-based robust filtering to handle outliers. The key properties of our approach are that \textit{(i)}~the learned biases generalize to different UWB anchor setups and \textit{(ii)} the approach is computationally efficient enough to run on resource-constrained hardware. We demonstrate our approach on a Crazyflie nano-quadcopter. Experimental results show that the proposed localization framework, relying only on the onboard IMU and UWB, provides an average of $42.08\%$ localization error reduction (in three different anchor setups) compared to the baseline approach without bias compensation. 
{We also show autonomous trajectory tracking on a quadcopter using our UWB TDOA localization approach.}

\end{abstract}

\section{Introduction and Related Work}
\label{sec:intro}

Over the last few decades, global navigation satellite systems (GNSS) have become an integral part of our daily lives, providing localization---under an open sky---with sub-meter accuracy anywhere on Earth~\cite{gps}. Today, indoor positioning solutions promise to enable similar capabilities for a plethora of indoor robotics applications (e.g., in warehouses, malls, airports, underground stations, etcetera).
Small and computationally-constrained indoor mobile robots have led researchers to pursue localization methods leveraging low-power and lightweight sensors. Ultra-wideband (UWB) technology, in particular, has been shown to provide sub-meter accurate, high-frequency, obstacle-penetrating ranging measurements that are robust to radio-frequency interference, using tiny integrated circuits~\cite{zafari2019survey}. UWB chips have already been included in the latest generations of smartphones~\cite{iphone} with the expectation that they will support faster data transfer and accurate indoor positioning, even in cluttered environments.  

\begin{figure}
  \centering
  \includegraphics[]{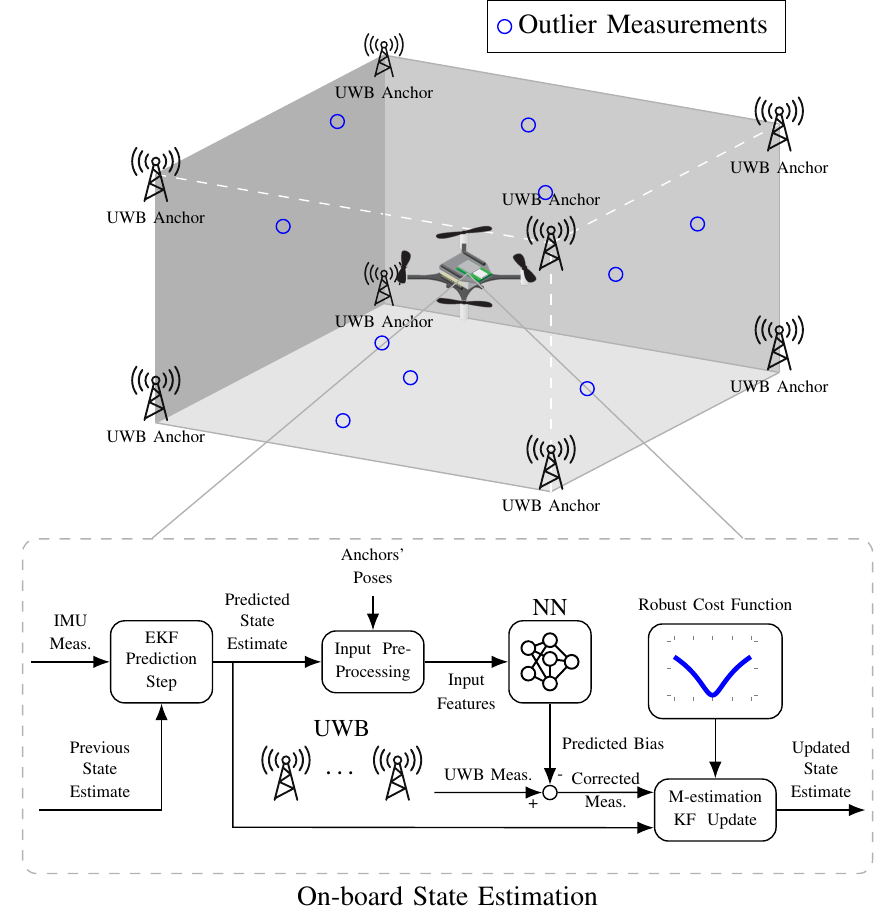}
  \caption{Sketch of our localization system setup (top) and block diagram of the proposed localization framework (bottom)---including the neural-network-based bias compensation and an M-estimation-based Kalman filter. Autonomous flight footage using the proposed localization scheme can be found at \url{http://tiny.cc/uwb-tdoa-bias-ral21}.}
  \label{fig:system}
\end{figure}

In autonomous indoor robotics~{\cite{cano2019kalman, kangultra20}}, the two main ranging schemes used for UWB localization are \textit{(i)} two-way ranging (TWR) and \textit{(ii)} time difference of arrival (TDOA).
The first is based on the time of flight (ToF) of a signal between an \emph{anchor} (a fixed UWB radio, Figure \ref{fig:system}) and a \emph{tag} (a mobile robot).
The second uses the difference between the arrival times of two signals---from different anchors---to one tag.
One of the perks of TDOA is that, unlike TWR, the number of required radio packets does not increase with the number of tracked robots/tags---as tags only passively listen to the messages exchanged between fixed UWB \emph{anchors}~\cite{kaplan2005understanding}. This enables TDOA localization to scale to a large number of devices, beyond what TWR could achieve.

Nonetheless, many factors can hinder the accuracy of UWB measurements, for either of the two schemes.
Non-line-of-sight (NLOS) and multi-path radio propagation, for example, can lead to erroneous, spurious measurements (so-called \emph{outliers}, Figure \ref{fig:system}). Even line-of-sight (LOS) UWB measurements exhibit error patterns (i.e.,  \emph{bias}), which are typically caused by the UWB antenna's radiation characteristic~\cite{tiemann2017ultra}. 
The ability to effectively \emph{(i)} remove outliers and \emph{(ii)} compensate for bias is essential to guarantee reliable and accurate UWB localization performance.

Multiple approaches have been proposed for the mitigation of UWB outlier measurements when using the TWR scheme. In~\cite{gururaj2017real}, a channel impulse response-based approach detects NLOS propagation from the received UWB waveforms, without the need for prior knowledge of the environment. In \cite{wymeersch2012machine}, the measurement error caused by NLOS is estimated directly from the received radio waveform using support vector machines (SVM) and Gaussian processes (GP).

Concerning the bias of TWR localization, the authors of ~\cite{ledergerber2017ultra} and~\cite{ledergerber2018calibrating} model and correct UWB pose-dependent biases using sparse pseudo-input Gaussian processes (SPGP) and demonstrate their approach on a quadrotor platform equipped with a Snapdragon Flight computer. 
{An iterative approach to estimate bias in TWR measurements is presented in \cite{van2020iterative}.}

As TDOA localization involves three UWB radios instead of two, modeling the measurement error is inherently more challenging. 
Most existing works focus on mitigating errors caused by NLOS and multi-path propagation.
Pioneering research was conducted in~\cite{prorok2014accurate,prorok2012online}, where an online expectation maximization (EM) algorithm addresses TDOA NLOS measurement errors. 
In~\cite{su2017semidefinite}, a semi-definite programming method is applied to the same problem. Much of the research on UWB TDOA localization including~\cite{prorok2014accurate,prorok2012online,su2017semidefinite} has been conducted in 2D scenarios and demonstrated using ground robots. In~\cite{hamer2018self}, the authors mention that UWB TDOA measurements are also affected by a systematic, position-dependent bias in line-of-sight conditions. Yet, to the best of our knowledge, no existing work focuses on addressing this spatially varying source of bias. 

In this work, we propose a framework to improve the accuracy and robustness of TDOA-based localization for resource-constrained mobile robots. We separately tackle the challenges posed by \textit{(i)} systematic bias and \textit{(ii)} outlier measurements. We leverage the representation power of neural networks (NN) to compensate for the bias. With the multi-radio nature of TDOA measurements in mind, we select appropriate input features to our NN model; in particular, we show that the bias is affected by the complete anchor pose and not just its position. Bypassing the need for raw UWB waveforms~\cite{wymeersch2012machine}, {we use M-estimation based Kalman filtering~\cite{chang2015huber} to handle outliers and improve localization robustness.}
We finally deploy our proposed approach on-board a Crazyflie~2.0 nano-quadcopter with limited computational resources.
We evaluate the proposed approach in flight experiments, and demonstrate the generalization capabilities of our approach by flying using three different, not previously seen UWB anchor setups.

Our main contributions can be summarized as follows:
\begin{enumerate}

  \item We propose a learning-based bias correction approach for UWB TDOA measurements, which generalizes to previously unobserved UWB anchor placements.
  
  \item  
  {We present a lightweight TDOA-based localization framework for resource-constrained mobile robots---combining bias correction and M-estimation\cite{chang2015huber}.}

  \item We implement the proposed framework on a nano-quadcopter and demonstrate the effectiveness and generalizability of our method by flying the nano-quadcopter using different UWB anchor setups. We show that our approach runs in real-time and in closed-loop on-board a nano-quadcopter yielding enhanced localization performance for autonomous trajectory tracking.
  
\end{enumerate}

{We use {localization} performance with the proposed M-estimation technique \emph{alone} as our baseline (note that a Crazyflie nano-quadcopter cannot {take off} reliably using the raw UWB TDOA measurements)}. 
Even compared to this baseline, the proposed localization framework achieves an average of  $42.08\%$ reduction in the root-mean-square (RMS) error of the position estimate for three previously unobserved UWB anchor constellations, providing an accuracy of approximately $0.14$m (RMS error). To the best of our knowledge, this work is the first demonstration of a lightweight UWB TDOA bias correction and robust localization framework on-board a nano-quadcopter for closed-loop flights.

 \section{UWB TDOA Measurements}
\label{sec:problem}

Our TDOA-based localization  setup is sketched in Figure~\ref{fig:system} (top). 
UWB localization systems can either rely on the time-of-flight of a signal---as in TWR---or the difference between the arrival times of two signals---as in TDOA---to compute range (w.r.t. one anchor) or range difference (w.r.t. two anchors), respectively. 
In TWR, two-way communication between an anchor and a tag is required to compute the range distance.
For TDOA, similar to a satellite positioning system, a set of stationary UWB anchors (whose positions are known) transmit radio signals into the surrounding space. Mobile robots equipped with UWB radio tags passively listen to these signals and localize themselves by comparing the arrival time of signals from each pair of anchors. Since the tags do not need active communications (unlike TWR), {TDOA-based localization systems scale better with the number of tags and are the more appropriate choice for {large-scale}, multi-robot applications}. To motivate our proposed approach---detailed in Section \ref{sec:DNN} and \ref{sec:localization_framework}---we start by analyzing some of the known limitations of existing UWB TDOA localization systems.

\begin{figure}
  \centering
  \includegraphics[]{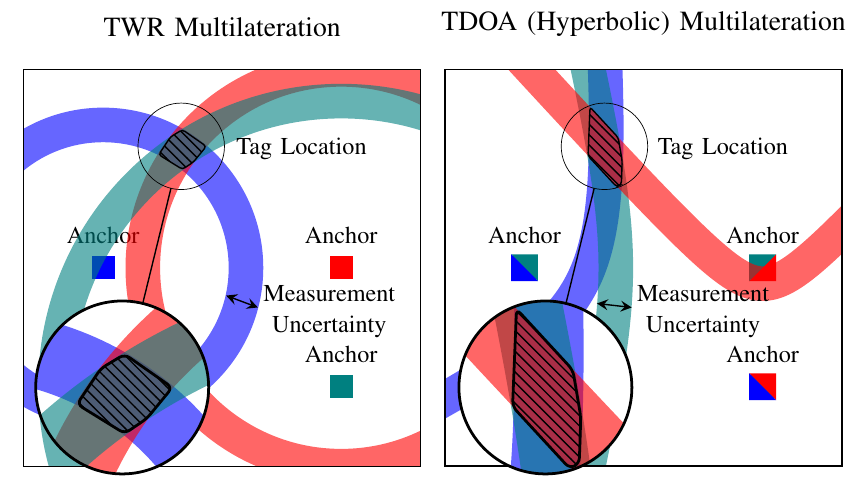}
  \caption{Two-dimensional comparison of TWR (left) and TDOA (right) multilateration 
  { of one tag using three anchors. 
  The same measurement uncertainty yields larger localization uncertainty for TDOA localization.} 
  }
\label{fig:hyperbolic}
\end{figure}

\subsection{Time Difference of Arrival Principles} 
\label{sec:toda_meas}

For TWR-based localization, two-way communication between anchor and tag is required to compute a range measurement. In the ideal case, in a TWR localization system with $m$ UWB anchors at positions $\bm{a}_i = [x_i,y_i,z_i]^T\in \mathbb{R}^3, i = 1,\dots,m$ and one tag at position $\mathbf{p} = [x,y,z]^T\in \mathbb{R}^3$, each of the $m$ range measurements $r_i$ can be written as:
\begin{equation}
  r_i = \|\mathbf{p}-\bm{a}_i\|,
  \label{eq:toa_measurement_model}
\end{equation}
where $\| \cdot \|$ is the Euclidean norm. TWR measurements can be used to solve the multilateration problem as the intersection of spheres {(see Figure~\ref{fig:hyperbolic})}.

In TDOA localization, the tag listens to messages from anchor $i$ and $j$ and compares the difference of arrival times of these two messages. The ideal TDOA measurement $d_{ij}$ can be written as:
\begin{equation}
  d_{ij} = \|\mathbf{p}-\bm{a}_i\| - \|\mathbf{p}-\bm{a}_j\|.
  \label{eq:TDOA_measurement_model}
\end{equation}
Geometrically, the \emph{locus of points} with a fixed distance difference from two given points (\emph{foci}) is a hyperbola {(Figure~\ref{fig:hyperbolic})}. 

However, in real world scenarios, UWB measurements (for both TWR and TDOA) are  corrupted by systematic errors, also called bias, and measurement noise. Therefore, a more realistic UWB TDOA measurement $\bar{d}_{ij}$ is:
\begin{equation}
\begin{split}
  \bar{d}_{ij} &= \|\mathbf{p}-\bm{a}_i\| - \|\mathbf{p}-\bm{a}_j\| + b_{ij}(\bm{\chi}) + n_{ij}\\
               &=d_{ij} + b_{ij}(\bm{\chi}) + n_{ij},
  \label{eq:realistic_measurement_model}
  \end{split}
\end{equation}
where $b_{ij}(\bm{\chi})$ indicates the systematic bias parametrized by a feature vector $\bm{\chi}$ and $n_{ij}\sim \mathcal{N}(0, \sigma^{2}_{ij})$ is a zero-mean Gaussian noise with variance $\sigma^{2}_{ij}$. Since hyperbolic localization {(Figure~\ref{fig:hyperbolic})} is more sensitive to imperfect measurements than TWR~\cite{sathyan2010analysis, wang2019doa}---especially outside or near the edges of the anchors' convex hull---compensating for the systematic bias is even more crucial for TDOA-based localization.

\begin{figure}
  \centering
  \includegraphics[]{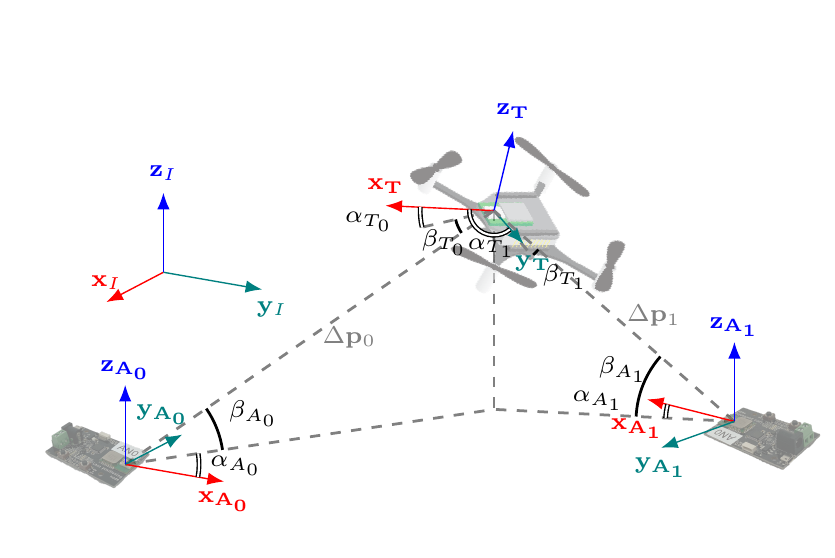}
  \vspace{-1em}
  \caption{ 
  Definition the relative poses between tag \emph{T} and anchors \emph{$A_0$}, \emph{$A_1$} through ranges ($\mathbf{\Delta \mathbf{p}}$'s), azimuth ($\alpha$'s), and elevation angles ($\beta$'s).
  }
  \label{fig:angle_define}
\end{figure}

\begin{figure*}
  \centering
  \includegraphics[]{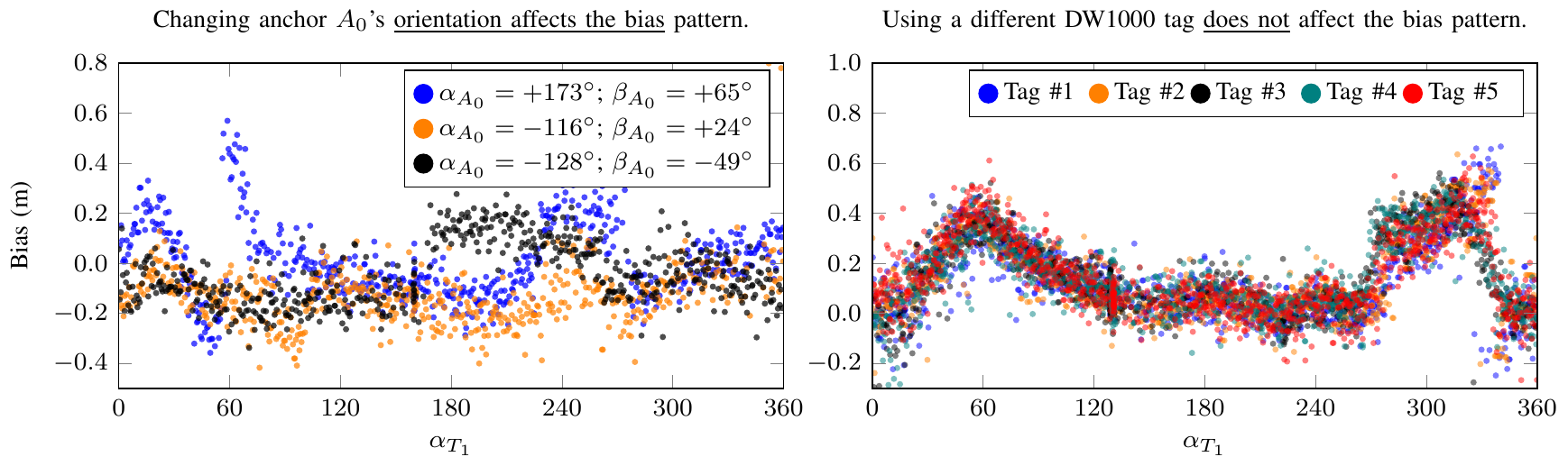}
  \caption{UWB TDOA measurement error for different orientations of anchor \emph{$A_0$} (left) and different DW1000 UWB tags (right). To generate the plot on the left, the orientation of \emph{$A_1$} was fixed at $\{\alpha_{A_1}: -160^\circ, \beta_{A_1}: 70^\circ\}$. On the right, the measurement errors for five different DW1000 tags when the poses of both anchors are fixed show a consistent pattern. See Figure~\ref{fig:angle_define} for the angle definitions.
}
  \vspace{-1.5em}
  \label{fig:TDOA_measurements}
\end{figure*}

\subsection{UWB TDOA Measurement Bias} 
\label{sec:uwb_singal_testing}

As reported in previous work on TWR localization~\cite{tiemann2017ultra, ledergerber2017ultra, van2020iterative}, off-the-shelf, low-cost UWB modules exhibit distinctive and reproducible error patterns.
TDOA measurements suffer from a similar systematic bias~\cite{hamer2018self}.

To demonstrate and characterize the TDOA bias in our experimental setup, we devised experiments using one tag and two Decawave DW1000 UWB anchors. First, we placed two DW1000 UWB anchors at a distance of $4.5$m from one another, in line-of-sight conditions. A Crazyflie nano-quadcopter mounted with a tag was commanded to spin around its z-axis while hovering at the midpoint between the two anchors. The TDOA measurements $\bar{d}_{01}$ from the tag was collected at $50$Hz. Ground truth values---for both UWB measurements and tag/anchor poses---were obtained through a millimeter-accurate motion capture system. 
We used the range-azimuth-elevation (RAE) model to uniquely define the relative pose between the tag and each of two anchors (see Figure~\ref{fig:angle_define}).
We repeated this experiment three times, each time changing the angles ($\alpha_{A_0},\beta_{A_0}$) of anchor $A_0$.

The TDOA measurement biases $b_{01}$ resulting from these experiments are presented in Figure \ref{fig:TDOA_measurements} (left). These measurements show that both the pose of the tag and the anchors have a non-negligible influence on the resulting bias pattern.
Furthermore, these biases proved consistent and reproducible through repeated experiments as they are ascribable to UWB radio's doughnut-shaped antenna pattern~\cite{rahayu2012radiation}.

In a second experiment, to assess the influence of the UWB chips' manufacturing variability, we repeated the experiment with five different DW1000 UWB tags, for fixed poses of the two anchors.
The resulting biases are shown in Figure \ref{fig:TDOA_measurements} (right). 
We note that the patterns created by the five different UWB chips are extremely similar to one another.
This is in contrast with the UWB TWR bias reportedly affected by small tag manufacturing differences in~\cite{ledergerber2017ultra, ledergerber2018calibrating}. 

{The results of the two experiments above suggest that the systematic bias in UWB TDOA measurements $b_{ij}(\bm{\chi})$ depends on the poses of the anchors and tags---i.e. they should appear in $\bm{\chi}$---but that it is consistent among different off-the-shelf DW1000 UWB tags---i.e. function $b_{ij}(\cdot)$ is the same for different tags.}

\subsection{UWB TDOA Outlier Measurements} 
\label{sec:NLOS}

Beyond the systematic biases observed in the previous section, TDOA measurements are often corrupted by outliers caused by multi-path and NLOS signal propagation. The multi-path effect is the result of the reflection of radio waves, leading to longer ToF and wrong TDOA measurements~\cite{steffes2012multipath}. In indoor scenarios, metal structures, walls, and obstacles are the major causes of multi-path propagation. NLOS propagation can occur because of the obstacle-penetrating capability of UWB radios, with delayed or degraded signals resulting also leading to outlier measurements~\cite{prorok2012online, su2017semidefinite}.

NLOS and multi-path propagation often result in extremely improbable TDOA measurements, which should be rejected as outliers. In Section~\ref{sec:localization_framework}, we devise a robust localization framework to reduce the influence of outliers and achieve reliable and robust localization performance.
 \section{UWB TDOA Bias Model Learning with a Neural Network}
\label{sec:DNN}

Knowing that TDOA hyperbolic localization is especially sensitive to measurement bias, we aim to show that compensating for the systematic bias can greatly improve the localization accuracy.

As highlighted in Section \ref{sec:uwb_singal_testing}, the TDOA systematic bias has a nonlinear pattern and is  dependent on the relative poses between UWB anchors and tags. Since two anchors and one tag are involved in each TDOA measurement (unlike one anchor and one tag in TWR), the TDOA systematic bias is the result of a complex relative-pose relationship between multiple UWB radios. We model this pose-dependent bias as a nonlinear function $b_{ij}(\Delta \mathbf{p},\bm{\alpha},\bm{\beta})$ of the relative positions  $\Delta \mathbf{p} = \left[\Delta \mathbf{p}_i^T, \Delta \mathbf{p}_j^T\right]$ with $\Delta \mathbf{p}_i = \left[x_i-x, y_i-y, z_i-z\right]^T$, relative azimuth angles $\bm{\alpha}=\left[\alpha_{A_i}, \alpha_{A_j}, \alpha_{T_i}, \alpha_{T_j}\right]^T$, and relative elevation angles $\bm{\beta} = \left[ \beta_{A_i}, \beta_{A_j}, \beta_{T_i}, \beta_{T_j}\right]^T$ between two anchors and the tag (see Figure~\ref{fig:angle_define}).
The UWB TDOA measurement model~\eqref{eq:realistic_measurement_model} can then be written as:
\begin{equation}
\begin{split}
  \bar{d}_{ij} &= d_{ij} + b_{ij}(\Delta \mathbf{p},\bm{\alpha},\bm{\beta}) + n_{ij}.
  \label{eq:biased_measurement_model}
  \end{split}
\end{equation}
In Section \ref{sec:feature_selection}, we further show the effectiveness and generalizability  of the bias model learned from these features (across previously unseen DW1000 tags and novel anchor placements)   compared to a model with a less rich input feature vector.

Since our work targets resource-constrained platforms, we propose to use a feed-forward neural network to learn a computationally efficient model for the complex UWB TDOA bias. 
For TWR bias compensation, an SPGP was previously proposed in~\cite{ledergerber2017ultra}. However, the time complexity of mean and variance prediction of an SPGP 
with $M$ pseudo-input points is $\mathcal{O}(M)$ and $\mathcal{O}(M^2)$, respectively~\cite{snelson2006sparse}. On resource-constrained platforms, the memory and power requirements of SPGP inference are often unattainable. To highlight this point, we compare the computational resources of the STM32F405 MCU used in many mobile robots (including the Crazyflie nano-quadcopter in our work) against those of \textit{(i)} the Odroid XU4 single board computer for quadcopters (used in~\cite{mathias2016autonomous}), and \textit{(ii)} the Qualcomm Snapdragon (used for TWR bias compensation in~\cite{ledergerber2017ultra}) in Table~\ref{tab:computation_resource}.

\begin{table}[b]
\caption{Computational Resources Comparison} 
\centering \begin{tabular}{c c c c} \toprule
    Name &
    STM32F405 &
    Odroid XU4 &
    Snapdragon Board \\
    \cmidrule(lr){1-4}
CPU &
    \makecell{1-core \\ 168MHz} &
    \makecell{ Exynos 5422 \\ Quad-core 2GHz} &
    \makecell{ Krait\\Quad-core 2.26GHz} \\ 
GPU &
    n/a &
    \makecell{Mali-T628 MP6}  &
    \makecell{Qualcomm Adreno 330} \\
DSP &
    n/a &
    n/a &
    Hexagon DSP \\
RAM  &
    196kB &
    2GB &
    2GB \\
\bottomrule
\end{tabular}
\label{tab:computation_resource} 
\end{table}

Both the Odroid XU4 and the Qualcomm Snapdragon board are equipped with powerful CPUs/GPUs and have large (2GB) system memories, making them much more suited for computationally intense tasks, even during flight. The STM32F405 has significantly less memory (196kB RAM) and a low-power CPU (1-core 168MHZ) and cannot run demanding SPGP-based bias compensation.

In contrast, the prediction time and memory requirements of a trained feed-forward neural network are fixed and only depend on the network architecture (rather than the amount of training data). Thus, the scalable (and potentially lower) computational and memory requirements make neural nets a fitting choice for resource-constrained platforms~\cite{duisterhoflearning}. 

Below, the localization framework using the neural network with the proposed input feature $\bm{\chi} = [\Delta \mathbf{p},\bm{\alpha},\bm{\beta}]^{T} \in \mathbb{R}^{14}$ is called ``proposed approach''. 
The output of the network is the predicted TDOA measurement bias $b_{ij}(\bm{\chi})\in\mathbb{R}$. We integrate the bias compensation into a Kalman filter (KF) framework for indoor localization. 
The architecture of the proposed network can be chosen to fit the computational limitations of the mobile robot platform.

In the results section, the proposed approach is compared to both \textit{(i)} a bare M-estimation-based EKF baseline (introduced in the next section) and \textit{(ii)} a NN-enhanced framework that \emph{does not} account for the anchors' orientations in $\bm{\chi}$.

The anchors' position and orientation are measured in advance using a Leica total station theodolite and stored on-board of the mobile robot. 
The details about data collection, the neural network architecture design, the training process, and the on-board implementation are provided in Sections~\ref{sec:NN_training} and \ref{sec:implementation}. In Section \ref{sec:experiments}, we also show that, with a fairly small network, we can model the impact of anchor-tag relative poses on the measurement bias, thus enabling real-time TDOA bias compensation on-board of a Crazyflie.

 \section{Localization Framework}
\label{sec:localization_framework}

In addition to pose-dependent bias, UWB TDOA localization is often plagued by outliers caused by unexpected NLOS and multi-path radio propagation. Unlike bias, these cannot be modeled without precise prior knowledge of the robots' trajectories and their surrounding environment. To reduce the influence of outliers, we use robust M-estimation.
Further, our approach can handle sparse UWB TDOA measurements, which can be a challenge for conventional Random Sample Consensus (RANSAC) approaches~\cite{mactavish2015all}.  A complete derivation of the M-estimation-based Kalman filter is beyond the scope of this paper.
While we provide the necessary equations for this paper, readers are referred to~\cite{chang2015huber} for further details.

\subsection{M-estimation-based Extended Kalman Filter} 
\label{sec:rkf}

For our UWB TDOA-based localization system, the system state $\mathbf{x}$ consists of the position $\mathbf{p}$, velocity $\mathbf{v}$, and the orientation of the tag. We first apply bias compensation to the TDOA measurements. After bias correction, the TDOA measurement w.r.t.~anchors $i$ and $j$ is given by:
\begin{equation}
    \begin{split}
     \label{eq:tdoa_after_biasnet}
     \tilde{d}_{ij} & = \bar{d}_{ij} - b_{ij}(\bm{\chi}) + n_{ij}\\
& = \|\mathbf{p}-\bm{a}_i\| - \|\mathbf{p}-\bm{a}_j\| + n_{ij}.
    \end{split}
\end{equation}
Since the tag only receives one TDOA measurement at a time, assuming the measurement noise is identically distributed for all anchor pairs, {the TDOA measurement $d_k$ at timestep $k$ can be written as:
\begin{equation}
    \label{eq:measurement_model}
    d_k = g(\mathbf{x}_k, n_k), \end{equation}
where $g(\cdot)$ is the TDOA measurement model and $n_k \sim \mathcal{N}(0, \sigma^{2})$ is the measurement noise.}
Then, we consider the following discrete-time, nonlinear system for TDOA localization (that is of general applicability to mobile robots): 
\begin{equation}
\label{eq:nonlinear_system}
  \begin{split}
    &\mathbf{x}_k = \mathbf{f}(\mathbf{x}_{k-1}, \mathbf{u}_k, \mathbf{w}_k), \\&d_k = g(\mathbf{x}_k, n_k), \end{split}
\end{equation}
where $\mathbf{x}_k \in \mathbb{R}^{N} $ is the system state at timestep $k$ with covariance matrix $\mathbf{P}_k\in\mathbb{R}^{N\times N}$, $\mathbf{f}(\cdot)$ is the motion model for a mobile robot with input $\mathbf{u}_k \in \mathbb{R}^{N}$ and process noise $\mathbf{w}_k  \sim \mathcal{N}(\mathbf{0}, \mathbf{Q}_k)$.

Due to the model nonlinearity, we use an M-estimation based extended Kalman filter (EKF) to estimate the states in \eqref{eq:nonlinear_system}. Replacing the quadratic cost function in a conventional Kalman filter with a robust cost function $\rho(\cdot)$---e.g. Geman-McClure (G-M), Huber or Cauchy~\cite{mactavish2015all}---we can write the posterior estimate as:
\begin{equation}
\label{eq:robustKF cost function}
  \hat{\mathbf{x}}_k = \argmin_{\mathbf{x}_k} \left( \sum_{i=1}^N \rho(e_{x,k,i}) +\rho(e_{d,k})\right),
\end{equation}
where $e_{d,k}=\frac{d_k-g(\mathbf{x}_k,0)}{\sigma}$ and $e_{x,k,i}$ is the element of:
\begin{equation}
\label{eq:robustfied_error}
\mathbf{e}_{x,k}(\mathbf{x}) = \mathbf{S}_k^{-1}(\mathbf{x}_k - \check{\mathbf{x}}_k) \\
\end{equation}
with prior estimates denoted as $\check{\mathbf{x}}_k$, and $\mathbf{S}_k$ being computed through the Cholesky factorization over the prior covariance matrix $\check{\mathbf{P}}_k$.

By introducing a weight function $w(e)\triangleq \frac{1}{e}\frac{\partial \rho(e)}{\partial e}$ for the process and measurement uncertainties---with $e\in \mathbb{R}$ as input---we can translate the optimization problem in \eqref{eq:robustKF cost function} into an Iterative Reweight Least-Square (IRLS) problem. Then, the optimal posterior estimate can be computed by iteratively solving the least-square problem using the robust weights computed from the previous solution.  

To initialize the iterative algorithm, we set $\hat{\mathbf{x}}_{k,0} = \check{\mathbf{x}}_k, \tilde{\mathbf{P}}_{k,0} = \check{\mathbf{P}}_k, \tilde{\sigma}^2_{k,0} = \sigma^2_k$. For brevity, we drop the timestep subscript $k$ in subsequent equations. In the $l$-th iteration, the rescaled covariance of the prior estimated state and the measurement can be written as:
\begin{equation}
\begin{split}
  \tilde{\mathbf{P}}_l &= \mathbf{S}_l \left(\mathbf{W}_{x,l}\right)^{-1} \left(\mathbf{S}_l\right)^{T},   \\
  \tilde{\sigma}^2_{l} & = \frac{\sigma^2_l}{w\left(e_{d,l}\right)},
\end{split}
\end{equation}
where $\mathbf{W}_{x,l}$  is the weighting matrix for process uncertainties with $w\left(e_{x,i,l}\right)$ in the diagonal entries, and $w\left(e_{d,l}\right)$ is the weight for the measurement uncertainty. Following the conventional EKF derivation, the weighted Kalman gain $\tilde{\mathbf{K}}_l$ is
\begin{equation}
  \tilde{\mathbf{K}}_l = \tilde{\mathbf{P}}_l\mathbf{G}_l^{T}\left(\mathbf{G}_l^{T}\tilde{\mathbf{P}}_l\mathbf{G}_l+\tilde{\sigma}^2_{l}\right)^{-1}, 
\end{equation}
where $\mathbf{G}_l$ is the Jacobian of the measurement model at $\hat{\mathbf{x}}_l$,
\begin{equation}
  \mathbf{G}_l = \left. \frac{\partial g(\mathbf{x}, 0)}{\partial \mathbf{x}} \right|_{\hat{\mathbf{x}}_l},
\end{equation}
The following IRLS iteration updates $\tilde{\mathbf{K}}_l$, $\tilde{\mathbf{P}}_l$, $\tilde{\sigma}^2_{l}$. For resource-constrained platforms, one can set a maximum number of iterations as a stopping criterion instead of convergence~\cite{chang2015huber}. After the final iteration $L$, the posterior state and covariance matrix can be computed as:
\begin{equation}
\begin{split}
    \hat{\mathbf{x}} &= \check{\mathbf{x}}_k + \tilde{\mathbf{K}}_L\left(d_k- g(\check{\mathbf{x}}_k, 0)\right), \\
    \hat{\mathbf{P}} &= \left(\mathbf{1} - \tilde{\mathbf{K}}_L\mathbf{G}_L\right)\tilde{\mathbf{P}}_L.
\end{split}
\end{equation}

\subsection{NN-Enhanced Robust Localization Framework}

Coupling the method in Section~\ref{sec:rkf} with the learning-based bias compensation proposed in Section~\ref{sec:DNN}, our overall localization framework (Figure~\ref{fig:system}, bottom) aims at improving both the accuracy and robustness of UWB TDOA localization. 
The on-board neural network corrects for bias before the M-estimation update step, thus making the measurement model in \eqref{eq:nonlinear_system} compliant with the zero-mean Gaussian distribution assumption.
Because of its general system formulation and the moderate computational requirements of both a pre-trained NN and the M-estimation-based EKF, the proposed neural network-enhanced TDOA-based localization framework is suitable for most resource-constrained platforms including mobile phones.

 \section{Experimental Results}
\label{sec:results}

To demonstrate the effectiveness and generalizability of the proposed localization framework, we implemented it onboard a Crazyflie 2.0 nano-quadcopter. We used eight UWB DW1000 modules from Bitcraze's Loco Positioning System (LPS) to set up the UWB TDOA localization system.
The ground truth position of the Crazyflie nano-quadcopter was provided by a motion capture system comprising of ten Vicon cameras. Note that the motion capture system is only used to collect training data and validate the localization performance. It is not required to set up the UWB localization system nor to fly the robot. The Crazyflie nano-quadcopter is equipped with a low-cost inertial measurement unit (IMU) and a UWB tag. 
All the software components of the proposed localization framework run onboard the Crazyflie microcontroller. {Footage of the autonomous flights is available at \url{http://tiny.cc/uwb-tdoa-bias-ral21}}.

\begin{figure}
  \centering
  \includegraphics[]{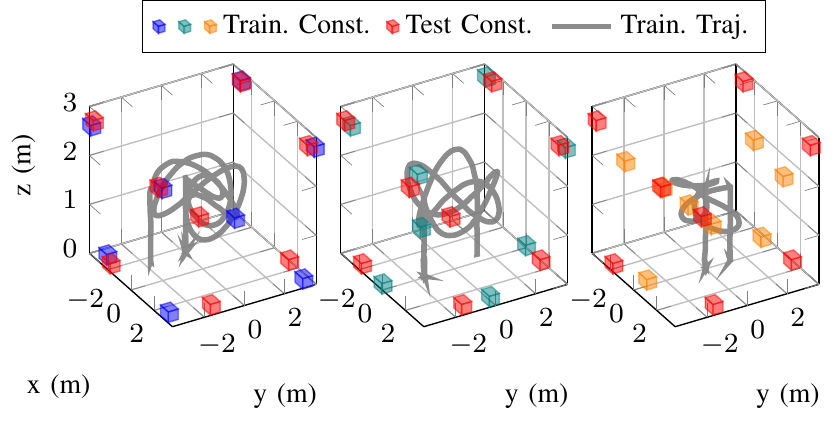}
  \caption{3-D plots of the UWB anchor constellations for data collection (train. const. $\#1, \#2, \#3$, from left to right) with examples of training trajectories.
Test constellation $\#1$ (used in Section \ref{sec:experiments}) is also overlaid.}
  \label{fig:training_constell}
\end{figure}

\subsection{Motion Model of the Nano-quadcopter}
The Crazyflie nano-quadcopter is modelled as a rigid body with double integrator dynamics, which is a simplified dynamic model for a quadcopter with an underlying position controller. The system is parameterized by a state $\mathbf{x}$ consisting of the nano-quadcopter's position $\mathbf{p}$, velocity $\mathbf{v}$, and orientation with respect to the inertial frame $\mathbf{C}_{IB}\in SO(3)$. {Under this simplified dynamic model, the system's state evolves as:
}
\begin{equation}
  \label{eq:ekf}
  \begin{split}
    & \dot{\mathbf{p}} = \mathbf{v},   ~~~
    \dot{\mathbf{v}} = \mathbf{C}_{IB}\bm{a} + \mathbf{g}, \\
    & \dot{\mathbf{C}}_{IB} = \mathbf{C}_{IB}\left[\bm{\omega}\right]_{\times},
  \end{split}
\end{equation}
where $\mathbf{g}$ is the gravitational acceleration, $\bm{a}\in \mathbb{R}^3$ and $\bm{\omega}\in \mathbb{R}^3$ are acceleration and the angular velocity in the body frame measured by the onboard IMU, and $\left[\cdot\right]_{\times}$ is the skew-symmetric operator defined as $\left[\bm{\omega}\right]_{\times}\mathbf{c} = \mathbf{c} \times \bm{\omega}, \forall \bm{\omega},\mathbf{c}\in \mathbb{R}^3$. Discretizing the dynamic model \eqref{eq:ekf} gives the motion model in \eqref{eq:nonlinear_system}, where IMU measurements are the system inputs.

After the \emph{(i)} EKF prediction and \emph{(ii)} NN bias compensation steps, we perform \emph{(iii)} M-estimation-based filtering using the G-M robust cost function.

\subsection{Data Collection and Network Training} 
\label{sec:NN_training}

To train our NN, we collected UWB TDOA measurements (and the associated ground truth labels) during a cumulative $\sim135$~minutes of real-world Crazyflie flights, using the three different UWB anchors setups (training constellations) shown in Figure \ref{fig:training_constell} {and different training trajectories {(products of multiple trigonometric functions whose amplitude, period and phase were randomized) to cover the indoor space.
We also varied yaw along the trajectory} to improve the representative power of our data set.}
We chose different constellations to represent a range of different geometries for our $7m\times8m\times3m$ flying arena.
The positions of anchors were measured using a total station with an accuracy of $5$mm root-mean-square (RMS) error, when compared to the motion capture system results. 
By measuring three non-coplanar points attached to the anchor at known positions, the orientations of the anchors were computed by point-cloud alignment \cite{barfoot2017state}, leading to azimuth and elevation angles within $1$ degree of the motion capture system measurements. 
Our dataset consists of over $800'000$ UWB measurements logged at 50~Hz {and is available at \url{http://tiny.cc/ral21-tdoa-dataset}}. From these, we subtracted the motion capture position information to compute the corresponding measurement error labels.
Measurement outliers caused by NLOS and multi-path effects with more than $1$m error were dropped from the dataset to focus on learning the antenna biases and not any outlier characteristics. 
Then, we partitioned this dataset into training, validation, and testing sets using a $70/15/15$ split.
The network was trained using \texttt{PyTorch}~\cite{paszke2017automatic} and halted when the error on the validation set increased over five consecutive iterations (early stopping) to prevent overfitting. As an optimizer, we chose mini-batch gradient descent~\cite{robbins1951stochastic}.
The testing set was used to evaluate the performance of the trained network.
The computing resources were provided by the Vector Institute.

\subsection{Implementation Onboard of a Nano-quadcopter} 
\label{sec:implementation}

The Crazyflie's limited memory is a major challenge for the on-board implementation of a sophisticated localization scheme. 
A Crazyflie 2.0 nano-quadcopter has $1$MB of flash storage and $196$kB of RAM, including $128$kB of static RAM and $64$kB of CCM (Core Coupled Memory). 
The default onboard firmware occupies $182$kB flash and $102$kB of static RAM are used by the basic estimation and control algorithms. 
To meet the memory constraints, we set our NN architecture to be a three-layer feed-forward network with $30$ neurons in each layer and fixed the number of iterations for the M-estimation update step to $2$. 
Both the NN and the M-estimation-based filter were implemented in plain \texttt{C}. The proposed localization framework software only takes approximately $12$kB of static RAM and $13$kB of flash storage, leaving $12\%$ and $81\%$ of the RAM and flash memory, respectively, free. 
We integrated this framework into the Crazyflie onboard EKF, running at $100$Hz.

\subsection{Input Features Selection Evaluation}
\label{sec:feature_selection}

The existing work in the literature~\cite{ledergerber2017ultra,ledergerber2018calibrating} does not propose to use the anchor orientations for TWR bias modeling.
Yet, as we showed in Section \ref{sec:uwb_singal_testing}, the anchor orientation has an impact on the systematic bias of TDOA measurements. In this subsection, we verify this observation by comparing the performance of a neural network trained with anchor orientations as part of its input features---our proposed approach---and one without them. Both networks have the same number of hidden layers and units, hyperparameters, and use the same training dataset.

\begin{figure}\centering
\includegraphics[]{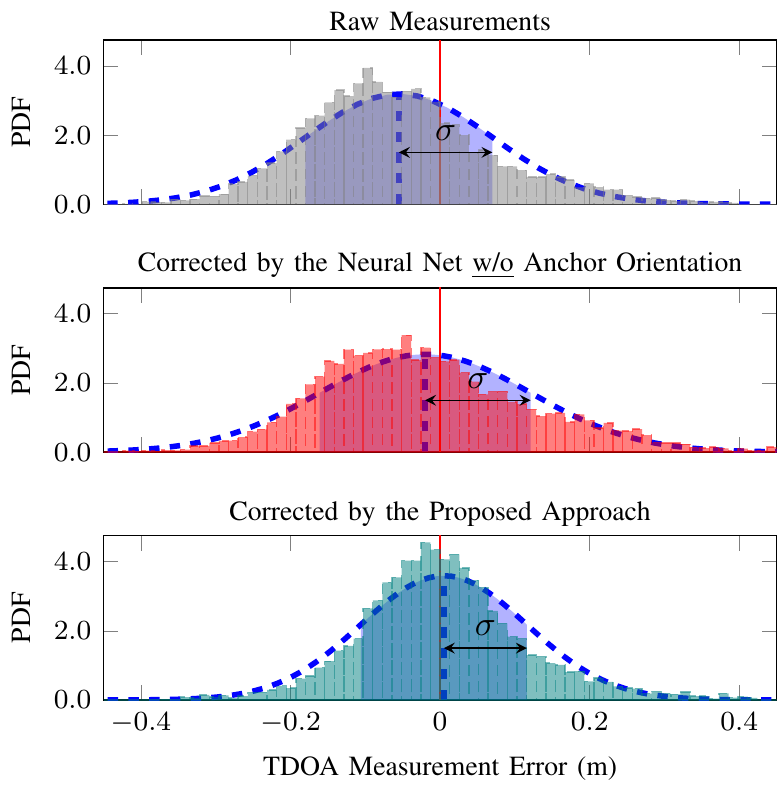}
  \caption{{Distribution of the TDOA measurement errors from Figure~\ref{fig:TDOA_measurements} (left) {before bias compensation} (grey), after correction with the NN \underline{not using anchor orientations} (red) and the proposed approach (teal). {Using the proposed approach (bottom plot), the fitted Gaussian probability density function (PDF), the blue dashed line, shows a lower bias and smaller standard deviation.}}
  }
  \label{fig:dnn_compare}
\end{figure}

{
In Figure~\ref{fig:dnn_compare} we present the normalized frequency distribution histograms of the TDOA measurement errors before and after bias correction (using the experimental data {from} Figure~\ref{fig:TDOA_measurements}).
Bias correction using the NN not having the anchor orientations as input improves the mean of the measurement error by {$3.5$~cm} (from {$-5.5$~cm} to {$-2.0$~cm}). However, the standard deviation $\sigma$ increases slightly, from {$12.5$~cm} to {$14.1$~cm}.
{Comparing to the NN without anchor orientation, the proposed approach provides $75.0\%$ and $21.3\%$ improvements in mean ($0.5$~cm) and the standard deviation ($11.1$~cm) of the measurement errors}, respectively, better matching a narrower (less uncertain) zero-mean Gaussian distribution.}

\subsection{Flight Experiments with Unseen Anchor Constellations} 
\label{sec:experiments}

To demonstrate the effectiveness of the proposed localization framework, we fly a Crazyflie nano-quadcopter using test anchor constellations that are different from those used for training (see Figure \ref{fig:training_constell}).
Without any of the components in our proposed localization framework, the Crazyflie quadcopter cannot reliably and repeatedly take off from the ground, due to the severe multi-path effect caused by the floor. 
With just the addition of the M-estimation-based EKF (to get rid of multi-path outlier measurements), the Crazyflie can take off and land. 
Therefore, we select the performance of the M-estimation-based EKF-only approach as our \emph{baseline} and compare it against \emph{(i)} the estimation enhanced through the proposed NN, and \emph{(ii)} the estimation enhanced through an NN without anchor orientations.
Both networks were trained using the process in Section~\ref{sec:NN_training}.

\begin{figure}
  \centering
  \includegraphics[]{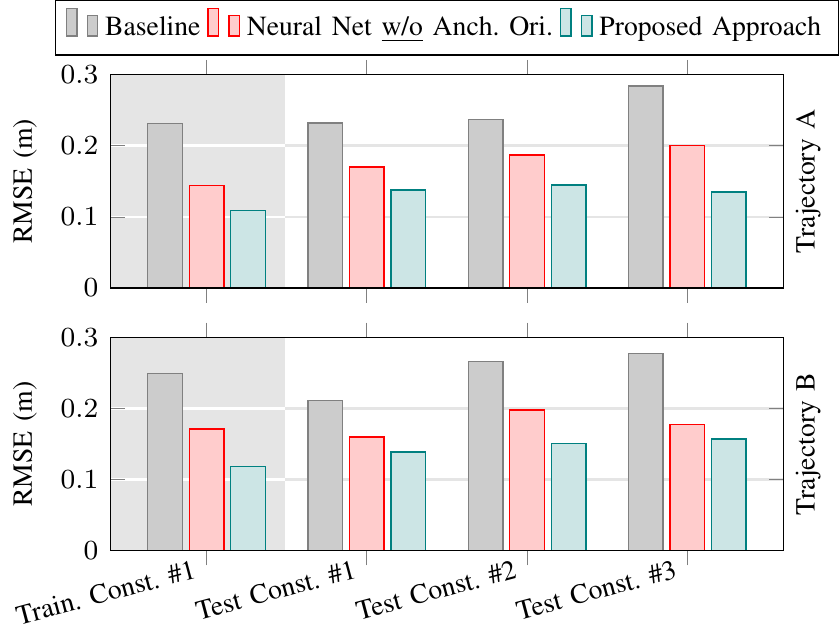}
  \caption{ 
  Root mean square (RMS) error of the nano-quadcopter position estimate with \emph{(i)} M-estimation based EKF-baseline (gray), \emph{(ii)} estimation enhanced through an NN without anchor orientations (red), and \emph{(iii)} estimation enhanced through the proposed NN approach (teal). Closed-loop flights on reference trajectories A and B were conducted using one training constellation and three different test constellations. The proposed localization framework \emph{(iii)} achieves $42.08\%$ and $20.32\%$ reduction of average RMS errors, w.r.t. \emph{(i)} and \emph{(ii)} in the test constellations.
  }
  \label{fig:rmse_barplot}
\end{figure}
We conducted flight experiments using training constellation $\#1$ and three entirely new anchor test constellations to show the generalization capability of the proposed localization framework. The Crazyflie nano-quadcopter was commanded to fly \emph{(i)} a planar (in x-y) circular trajectory, called trajectory A, and \emph{(ii)} a circular (in x-y) trajectory with sinusoidal height (i.e. with varying z position), called trajectory B. Neither of these two trajectories was among the training trajectories. The RMS errors of the three localization methods (M-estimation-based EKF alone, EKF plus bias correction with the proposed approach, or EKF plus NN without anchor orientations) are summarized in Figure~\ref{fig:rmse_barplot}.

\begin{figure}
  \centering
  \includegraphics[]{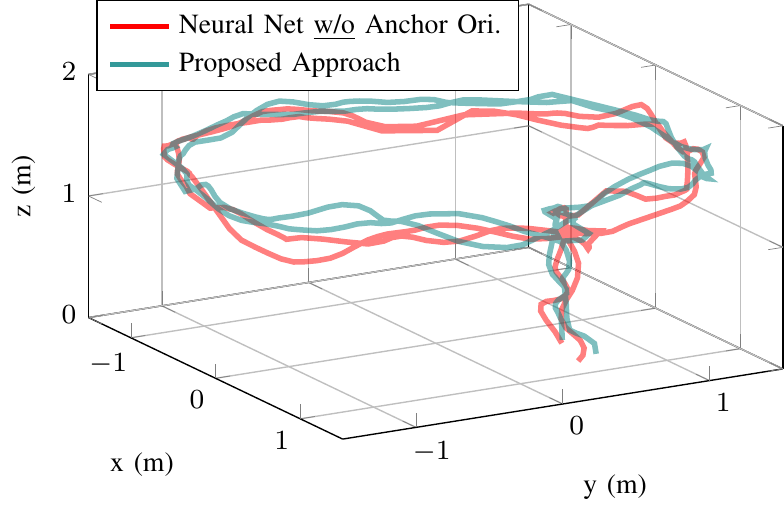}
  \vspace{-0.5em}
  \caption{Closed-loop flight paths for reference trajectory A with estimation enhanced through an NN without anchor orientations (red) and estimation enhanced through the proposed approach (teal), in test constellation $\#1$.}
  \label{fig:testing_contell_traj}
\end{figure}

In the training constellation setup, the proposed method provides $52.71\%$ and $27.65\%$ average RMS error reduction comparing to \emph{(i)} M-estimation-based EKF alone and \emph{(ii)} M-estimation-based EKF enhanced through an NN without anchor orientations. In the three test constellations, the proposed framework achieves $42.08\%$ and $20.32\%$ reduction of average RMS errors, w.r.t. \emph{(i)} and \emph{(ii)}, leading to an accuracy of approximately $0.14$m RMS localization error on-board a Crazyflie nano-quadcopter.
We demonstrate the on-board estimation results of trajectory A (see Figure \ref{fig:testing_contell_traj}) using test anchor constellation $\#1$ as an example.

\begin{figure*}
  \centering
  \includegraphics[]{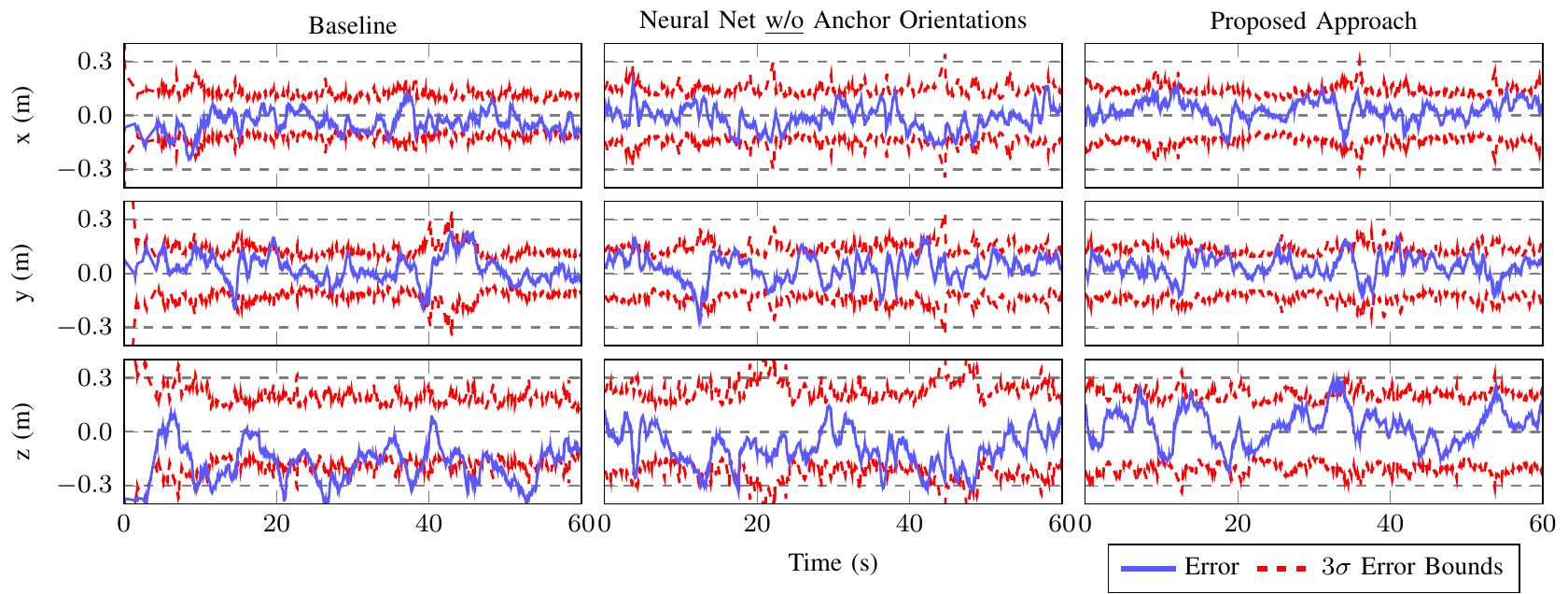}
  \caption{ 
  Estimation error (blue) and 3-$\sigma$ error bounds (red) for reference trajectory A using test const. $\#1$. Without bias compensation, the error can exceed the 3-$\sigma$ bound (e.g. in z). 
  Compared to the NN without anchor orientations, the proposed NN improves and debiases estimation along the z-axis. 
  }
  \vspace{-1.5em}
\label{fig:estimation_results}
\end{figure*}

The onboard estimation errors computed with respect to the ground truth and the estimated three-$\sigma$ uncertainty bounds for the three approaches during the closed-loop flight are shown in Figure~\ref{fig:estimation_results}.
Both NN-enhanced approaches show a reduction of the RMS estimation error compared to the baseline (M-estimation-based EKF), especially along the z-axis.
The larger estimation error in the z direction, before bias correction, can be partly attributed to our setup's specific geometry, having a narrower anchor separation in z ($\sim$2.8 meters).
We also observe that, with the M-estimation based EKF alone, the estimation errors are out of the estimated three-$\sigma$ error bounds. This phenomenon is caused by the uncompensated UWB measurement biases. With biased measurements, KF-type estimators will provide biased estimates and overconfident uncertainty, leading to inconsistent estimation results~\cite{barfoot2017state}. With NN bias compensation, most of the estimation errors are within three-$\sigma$ error bounds. Also, compared to the NN without anchor orientations, the proposed approach provides an improved and unbiased estimation along the z-axis.

 \section{Conclusions}
\label{sec:conclusions}

In this article, we presented a learning-enhanced, robust TDOA-based localization framework for resource-constrained mobile robots. 
To compensate for the systematic biases in the TDOA measurements, we proposed a lightweight neural network model and selected appropriate input features based on the analysis of the TDOA measurement error patterns. 
For robustness, we used the M-estimation technique to down-weight outliers. 
The proposed localization framework is frugal enough to be implemented on-board a Crazyflie 2.0 nano-quadcopter. 
We demonstrated the effectiveness and generalizability of our approach through real-world flight experiments---using multiple different anchor test constellations. Experimental results show that the proposed approach provides an average of $42.09\%$ localization error reduction compared to the baseline method without bias compensation.
In summary, our approach \emph{(i)} allows for real-time execution on-board a nano-quadcopter during flight, \emph{(ii)} yields enhanced localization performance for autonomous trajectory tracking, and \emph{(iii)} generalizes to previously unobserved UWB anchor constellations.

\section{Acknowledgments}
\label{sec:acknowledgments}

We would like to thank Bardienus P. Duisterhof (Delft University of Technology) for the insights regarding the on-chip neural network implementation and Kristoffer Richardsson (Bitcraze) for his guidance on embedded coding.


\begin{thebibliography}{10}
\providecommand{\url}[1]{#1}
\csname url@rmstyle\endcsname
\providecommand{\newblock}{\relax}
\providecommand{\bibinfo}[2]{#2}
\providecommand\BIBentrySTDinterwordspacing{\spaceskip=0pt\relax}
\providecommand\BIBentryALTinterwordstretchfactor{4}
\providecommand\BIBentryALTinterwordspacing{\spaceskip=\fontdimen2\font plus
\BIBentryALTinterwordstretchfactor\fontdimen3\font minus
  \fontdimen4\font\relax}
\providecommand\BIBforeignlanguage[2]{{\expandafter\ifx\csname l@#1\endcsname\relax
\typeout{** WARNING: IEEEtran.bst: No hyphenation pattern has been}\typeout{** loaded for the language `#1'. Using the pattern for}\typeout{** the default language instead.}\else
\language=\csname l@#1\endcsname
\fi
#2}}

\bibitem{gps}
\BIBentryALTinterwordspacing
GPS.gov. (2017) {GPS} accuracy. [Online]. Available:
  \url{https://www.gps.gov/systems/gps/performance/accuracy/#how-accurate}
\BIBentrySTDinterwordspacing

\bibitem{zafari2019survey}
F.~Zafari, A.~Gkelias, and K.~K. Leung, ``A survey of indoor localization
  systems and technologies,'' \emph{IEEE Communications Surveys \& Tutorials}, 2019,
  vol.~21, no.~3, pp. 2568--2599.

\bibitem{iphone}
\BIBentryALTinterwordspacing
B.~Barrett. (2019) The biggest {iPhone} news is a tiny new chip inside it.
  [Online]. Available: \url{https://www.wired.com/story/apple-u1-chip/}
\BIBentrySTDinterwordspacing

\bibitem{cano2019kalman}
J.~Cano, S.~Chidami, and J.~Le~Ny, ``A Kalman filter-based algorithm for
  simultaneous time synchronization and localization in UWB networks,'' in
  \emph{IEEE International Conference on Robotics and Automation (ICRA)}, 2019,  pp. 1431--1437.

\bibitem{kangultra20}
J.~Kang, K.~Park, Z.~Arjmandi, G.~Sohn, M.~Shahbazi, and P.~M{\'e}nard,
  ``Ultra-wideband aided {UAV} positioning using incremental smoothing with
  ranges and multilateration,'' in \emph{IEEE/RSJ International Conference on Intelligent Robots and Systems (IROS)}, 2020, pp. 4529--4536.


\bibitem{kaplan2005understanding}
E.~Kaplan and C.~Hegarty, \emph{Understanding GPS: principles and
  applications}.\hskip 1em plus 0.5em minus 0.4em\relax Artech House, 2005.

\bibitem{tiemann2017ultra}
J.~Tiemann, J.~Pillmann, and C.~Wietfeld, ``Ultra-wideband antenna-induced
  error prediction using deep learning on channel response data,'' in
  \emph{IEEE 85th Vehicular Technology Conference (VTC Spring)}, 2017, pp. 1--5.

\bibitem{gururaj2017real}
K.~Gururaj, A.~K. Rajendra, Y.~Song, C.~L. Law, and G.~Cai, ``Real-time
  identification of {NLOS} range measurements for enhanced {UWB}
  localization,'' in \emph{International Conference on Indoor
  Positioning and Indoor Navigation (IPIN)}, 2017, pp. 1--7.

\bibitem{wymeersch2012machine}
H.~Wymeersch, S.~Maran{\`o}, W.~M. Gifford, and M.~Z. Win, ``A machine learning approach to ranging error mitigation for UWB localization,'' \emph{IEEE Trans. Commun.}, 2012, vol.~60, no.~6, pp. 1719--1728.

\bibitem{ledergerber2017ultra}
A.~Ledergerber and R.~D'Andrea, ``Ultra-wideband range measurement model with
  {Gaussian} processes,'' in \emph{IEEE Conference on Control
  Technology and Applications (CCTA)}, 2017, pp. 1929--1934.

\bibitem{ledergerber2018calibrating}
A.~Ledergerber and R.~D’andrea, ``Calibrating away inaccuracies in ultra
  wideband range measurements: A maximum likelihood approach,'' \emph{IEEE
  Access}, 2018, vol.~6, pp. 78\,719--78\,730.

\bibitem{van2020iterative}
B.~van~der Heijden, A.~Ledergerber, R.~Gill, and R.~D’Andrea, ``Iterative
  bias estimation for an ultra-wideband localization system,'' in \emph{International Federation of Automatic Control (IFAC)}, 2020.

\bibitem{prorok2014accurate}
A.~Prorok and A.~Martinoli, ``Accurate indoor localization with ultra-wideband
  using spatial models and collaboration,'' \emph{The International Journal of
  Robotics Research}, 2014, vol.~33, no.~4, pp. 547--568.

\bibitem{prorok2012online}
A.~Prorok, L.~Gonon, and A.~Martinoli, ``Online model estimation of
  ultra-wideband {TDOA} measurements for mobile robot localization,'' in
  \emph{IEEE International Conference on Robotics and Automation
  (ICRA)}, 2012, pp. 807--814.

\bibitem{su2017semidefinite}
Z.~Su, G.~Shao, and H.~Liu, ``Semidefinite programming for {NLOS} error
  mitigation in {TDOA} localization,'' \emph{IEEE Communications Letters}, 2017, vol.~22, no.~7, pp. 1430--1433.

\bibitem{hamer2018self}
M.~Hamer and R.~D’Andrea, ``Self-calibrating ultra-wideband network
  supporting multi-robot localization,'' \emph{IEEE Access}, 2018, vol.~6, pp.
  22\,292--22\,304.

\bibitem{chang2015huber}
L.~Chang, K.~Li, and B.~Hu, ``Huber’s {M}-estimation-based process
  uncertainty robust filter for integrated {INS/GPS},'' \emph{IEEE Sensors
  Journal}, 2015, vol.~15, no.~6, pp. 3367--3374.

\bibitem{sathyan2010analysis}
T.~Sathyan, M.~Hedley, and M.~Mallick, ``An analysis of the error
  characteristics of two time of arrival localization techniques,'' in
  \emph{IEEE 13th International Conference on Information Fusion}, 2010, pp. 1--7.

\bibitem{wang2019doa}
R.~Wang, Z.~Chen, and F.~Yin, ``{DOA}-based three-dimensional node geometry
  calibration in acoustic sensor networks and its cram{\'e}r--rao bound and
  sensitivity analysis,'' \emph{IEEE/ACM Trans. Audio, Speech, Language
  Process}, 2019, vol.~27, no.~9, pp. 1455--1468.


\bibitem{rahayu2012radiation}
Y.~Rahayu and R.~Ngah, ``Radiation pattern characteristics of multiple
  band-notched ultra wideband antenna with asymmetry slot reconfiguration,''
  \emph{International Journal on Advances in Telecommunications}, 2012, vol. 5,
  no.~1\&2.

\bibitem{steffes2012multipath}
C.~Steffes and S.~Rau, ``Multipath detection in {TDOA} localization
  scenarios,'' in \emph{IEEE Workshop on Sensor Data Fusion: Trends, Solutions, Applications (SDF)}, 2012, pp. 88--92.

\bibitem{snelson2006sparse}
E.~Snelson and Z.~Ghahramani, ``Sparse gaussian processes using
  pseudo-inputs,'' in \emph{Advances in Neural Information Processing Systems (NIPS)},
  2006, pp. 1257--1264.

\bibitem{mathias2016autonomous}
H.~D. Mathias, ``An autonomous drone platform for student research projects,'' \emph{Journal of Computing Sciences in Colleges}, 2016, vol.~31, no.~5, pp. 12--20.

\bibitem{duisterhoflearning}
B.~P. Duisterhof, S.~Krishnan, J.~J. Cruz, C.~R. Banbury, W.~Fu, A.~Faust,
  G.~C. de~Croon, and V.~J. Reddi, ``Learning to seek: Autonomous source
  seeking with deep reinforcement learning onboard a nano drone
  microcontroller,'' \emph{arXiv preprint arXiv:1909.11236}, 2019.

\bibitem{mactavish2015all}
K.~MacTavish and T.~D. Barfoot, ``At all costs: A comparison of robust cost
  functions for camera correspondence outliers,'' in \emph{IEEE Conference on Computer and Robot Vision (CRV)}. 2015, pp. 62--69.

\bibitem{barfoot2017state}
T.~D. Barfoot, \emph{State estimation for robotics}.\hskip 1em plus 0.5em minus 0.4em\relax Cambridge University Press, 2017.

\bibitem{paszke2017automatic}
A.~Paszke, S.~Gross, S.~Chintala, G.~Chanan, E.~Yang, Z.~DeVito, Z.~Lin,
  A.~Desmaison, L.~Antiga, and A.~Lerer, ``Automatic differentiation in
  {PyTorch},'' 2017.

\bibitem{robbins1951stochastic}
H.~Robbins and S.~Monro, ``A stochastic approximation method,'' \emph{The
  Annals of Mathematical Statistics}, 1951, pp. 400--407.

\end{thebibliography}
\end{document}